%% file: 0.main.tex
\pdfoutput=1

\PassOptionsToPackage{table}{xcolor}

\documentclass[11pt]{article}

\usepackage[preprint]{acl}

\usepackage{times}
\usepackage{latexsym}

\usepackage[T1]{fontenc}

\usepackage[utf8]{inputenc}
\usepackage{microtype}
\usepackage{inconsolata}
\usepackage{graphicx}
\usepackage{url}            
\usepackage{booktabs}       
\usepackage{amsfonts}       
\usepackage{nicefrac}       
\usepackage{microtype}      
\usepackage{xspace}
\usepackage{soul}
\usepackage{caption}
\usepackage{siunitx}
\usepackage{bbm}
\usepackage{subcaption}

\usepackage{tabularx}
\definecolor{mydarkblue}{RGB}{0, 0, 139}
\definecolor{whitegray}{RGB}{243, 243, 243}
\definecolor{myorange}{HTML}{ED553B}
\definecolor{mymint}{HTML}{3CAEA3}
\definecolor{myblue}{HTML}{20639B}
\definecolor{mynavy}{HTML}{173F5F}
\definecolor{myyellow}{HTML}{F6D55C}

\hypersetup{
    colorlinks=true,
    linkcolor=mydarkblue,
    filecolor=mydarkblue,
    citecolor=mydarkblue,
    urlcolor=mydarkblue,
}
\usepackage{gensymb}
\usepackage{mathtools}
\usepackage{bm}
\usepackage[capitalize, noabbrev, nameinlink]{cleveref}
\usepackage{algorithm}
\usepackage{enumitem}
\usepackage{wrapfig}
\usepackage{pifont}
\usepackage{multirow}
\usepackage{tabularx}
\usepackage{hyperref}
\usepackage{mathtools}
\usepackage{wasysym}
\usepackage{booktabs}
\usepackage{tcolorbox}
\usepackage{listings}

\newcommand{\Algname}{$\textsc{CleanMol}$\xspace}

\newcommand{\cll}{\cellcolor{blue!4}}

\title{Improving Chemical Understanding of LLMs via SMILES Parsing}

\author{Yunhui Jang \\
 KAIST \\
  \texttt{yunhuijang@kaist.ac.kr} \\\And
  Jaehyung Kim \\
  Yonsei University\\
  \texttt{jaehyungk@yonsei.ac.kr} \\ \\\And
  Sungsoo Ahn \\
    KAIST \\
    \texttt{sungsoo.ahn@kaist.ac.kr}}

\begin{document}
\maketitle
\begin{abstract}
Large language models (LLMs) are increasingly recognized as powerful tools for scientific discovery, particularly in molecular science. A fundamental requirement for these models is the ability to accurately understand molecular structures, commonly encoded in the SMILES representation. However, current LLMs struggle to interpret SMILES, even failing to carry out basic tasks such as counting molecular rings. To address this limitation, we introduce \Algname{}, a novel framework that formulates SMILES parsing into a suite of clean and deterministic tasks explicitly designed to promote graph-level molecular comprehension. These tasks span from subgraph matching to global graph matching, providing structured supervision aligned with molecular structural properties. We construct a molecular pretraining dataset with adaptive difficulty scoring and pre-train open-source LLMs on these tasks. Our results show that \Algname{} not only enhances structural comprehension but also achieves the best or competes with the baseline on the Mol-Instructions benchmark.

\end{abstract}

\input{1.introduction}

\input{2.motivation_bottleneck}
\input{3.method}

\input{4-1.experiments_fundamental}

\input{4.experiments}

\input{5.related}
\input{5.conclusion}

\clearpage

\section*{Broader Impact}
Our work contributes to the development of structurally grounded models for molecular applications. By introducing a structured, clean, and scalable set of SMILES parsing tasks, we aim to equip LLMs with a stronger inductive bias toward molecular structure understanding. This can enhance downstream applications such as drug discovery, materials design, and reaction prediction by improving the fidelity and reliability of molecular reasoning. However, as with any generative AI system in chemistry, potential misuse remains a concern. The capacity to generate toxic, harmful, or restricted compounds necessitates careful integration of safety measures and expert oversight.

\section*{Limitations}
\paragraph{Limited structural information.}
Our SMILES parsing tasks focus on graph-level molecular structures and do not incorporate 3D conformational information, which is essential for many biological and physicochemical applications. Additionally, while our tasks are deterministic and scalable, they do not capture more nuanced chemical features such as stereochemistry, electronic effects, or reactivity patterns, which often require context beyond 2D topological graphs.

\paragraph{Language-specific scope.}
Our experiments are conducted exclusively in English and do not explore the applicability of the method across other languages, including morphologically rich or typologically diverse ones. Given that behaviors can vary across languages due to linguistic structure and training data distributions, the generalizability of our approach to multilingual settings remains an open question.

\paragraph{Model and dataset scale.}
Due to computational constraints, our experiments are limited to language models with up to 7.5B–8B parameters. It remains to be seen whether our framework scales effectively to larger models (e.g., 70B or beyond). Moreover, our pretraining is performed on a relatively modest dataset of 250K molecules, and while we observe consistent improvements, further studies on larger-scale datasets are necessary to assess the robustness and scalability of the approach.


\bibliography{custom}

\clearpage
\appendix

\input{8.appendix}

\end{document}

%% file: 1.introduction.tex
\section{Introduction}

Molecular string representations such as SMILES~\citep{weininger1988smiles} and SELFIES~\citep{krenn2020self} have become a standard format for applying large language models (LLMs) to chemistry. These one-dimensional strings flatten molecular graphs by traversing atoms and bonds and are syntactically compatible with LLMs~\citep{xia2025naturelanguagemodeldeciphering, taylor2022galactica, edwards2022molt5, christofidellis2023chemt5, pei-etal-2023-biot5, fang2024molinstructions}. As a result, most molecular LLMs adopt training paradigms from the natural language processing domain, treating molecular strings as sequences of tokens analogous to sentences in natural language.

However, molecular strings follow complex syntactic rules for encoding molecular structures, which LLMs often struggle to interpret. For instance, SMILES grammar includes specific conventions to denote rings and branches—often involving non-contiguous tokens to represent connected substructures. Additionally, SMILES representations must satisfy structural constraints such as proper valency and ring closure. As a result, current LLMs often misinterpret SMILES, which implies a failure to capture the underlying molecule represented by the SMILES string. This is reflected in their inability to perform even basic tasks, such as counting the number of rings or producing consistent outputs for different SMILES strings of the same molecule~\citep{jang2024chain, white2023assessment, ganeeva2024lost}. Our experiments revisit such limitations, as shown in \cref{fig:1_parsing} and \cref{subsec:2_failure_smiles_parsing}.

\input{figure/fig_2_failure}

One might expect such an understanding would ``naturally emerge'' from training LLMs on large corpora of SMILES strings for downstream tasks such as molecular generation and retrosynthetic analysis. However, high-quality data is limited and difficult to obtain. Unlike text or image data, which can be gathered at scale via web scrapping, chemical data often require expensive wet lab experiments or simulations for annotation. Although open-source datasets such as USPTO series~\citep{wei2010novel, lu2022unified} and MoleculeNet~\citep{wu2018moleculenet} exist, their scale remains modest compared to datasets in other domains~\citep{deng2009imagenet, raffel2020exploring, lozhkov2024starcoder}. Consequently, most chemical LLMs often rely on ambiguous and indirect pretraining objectives with non-deterministic and unclear tasks (e.g., masking each token in SMILES and reconstruct them or translation between a molecular string and its description) \citep{pei-etal-2023-biot5, edwards2022molt5}, or focus on instruction tuning with limited-scale datasets~\citep{fang2024molinstructions, yu2024llasmol}.

In response, we propose \textit{SMILES parsing}—a suite of clean, deterministic, and scalable tasks that require models to extract structural information from molecular strings, as illustrated in \cref{fig:1_parsing}. We argue that a natural and necessary candidate task for training LLMs to understand the SMILES representation is the extraction of deterministic graph-level information from molecular structures. To address this, we define five SMILES parsing tasks including subgraph matching (e.g., functional group, ring size, and chain length) and global graph matching (e.g., SMILES canonicalization and fragment assembly). Each task provides unambiguous supervision with deterministic answers. Based on these tasks, we construct the \Algname dataset, consisting of 250K molecules annotated via lightweight molecular graph analysis tools such as RDKit~\citep{greg2024rdkit}. Notably, our approach is scalable since the annotations for these tasks do not require any experiment or human annotation, in principle, SMILES parsing can be applied to all the existing molecules in the real world.

To evaluate and demonstrate the benefit of our new \Algname dataset, we also introduce a two-stage training framework: first, the model is pre-trained on the proposed SMILES parsing tasks and then fine-tuned on downstream chemical applications. To enhance data efficiency in the first stage, we propose a task-adaptive data pruning that selects structurally informative molecules and a curriculum learning framework that organizes them from easy to hard order.

We empirically validate our approach by training recent LLM backbones~\citep{grattafiori2024llama, yang2024qwen2} and evaluating them on three downstream tasks from the Mol-Instructions benchmark~\citep{fang2024molinstructions}, including retrosynthesis, reagent prediction, and forward reaction prediction. Surprisingly, our clean and structure-aware \Algname{} framework enables the models to achieve state-of-the-art or competitive results on the downstream tasks. This demonstrates that incorporating deterministic structural supervision via SMILES parsing can significantly enhance molecular generation capabilities, even without direct exposure to generation-specific training data.

We summarize our contributions as follows:

\begin{itemize}
    \item We revisit the limitations of LLMs in interpreting molecular strings, highlighting the structural bottleneck.
    \item We propose five deterministic and scalable SMILES parsing tasks and introduce the \Algname dataset to bridge the gap between string-level and graph-level molecular understanding of LLMs.
    \item We design a two-stage training framework incoporating a task-adaptive data pruning and curriculum learning strategy.
    \item We validate the impact of \Algname by demonstrating a consistent performance improvement across multiple downstream tasks.
\end{itemize}

%% file: figure/fig_2_failure.tex
\begin{figure*}[t]
    \vspace{-0.1in}
\centering
    \begin{subfigure}[t]{0.64\textwidth}
    \includegraphics[width=\linewidth]{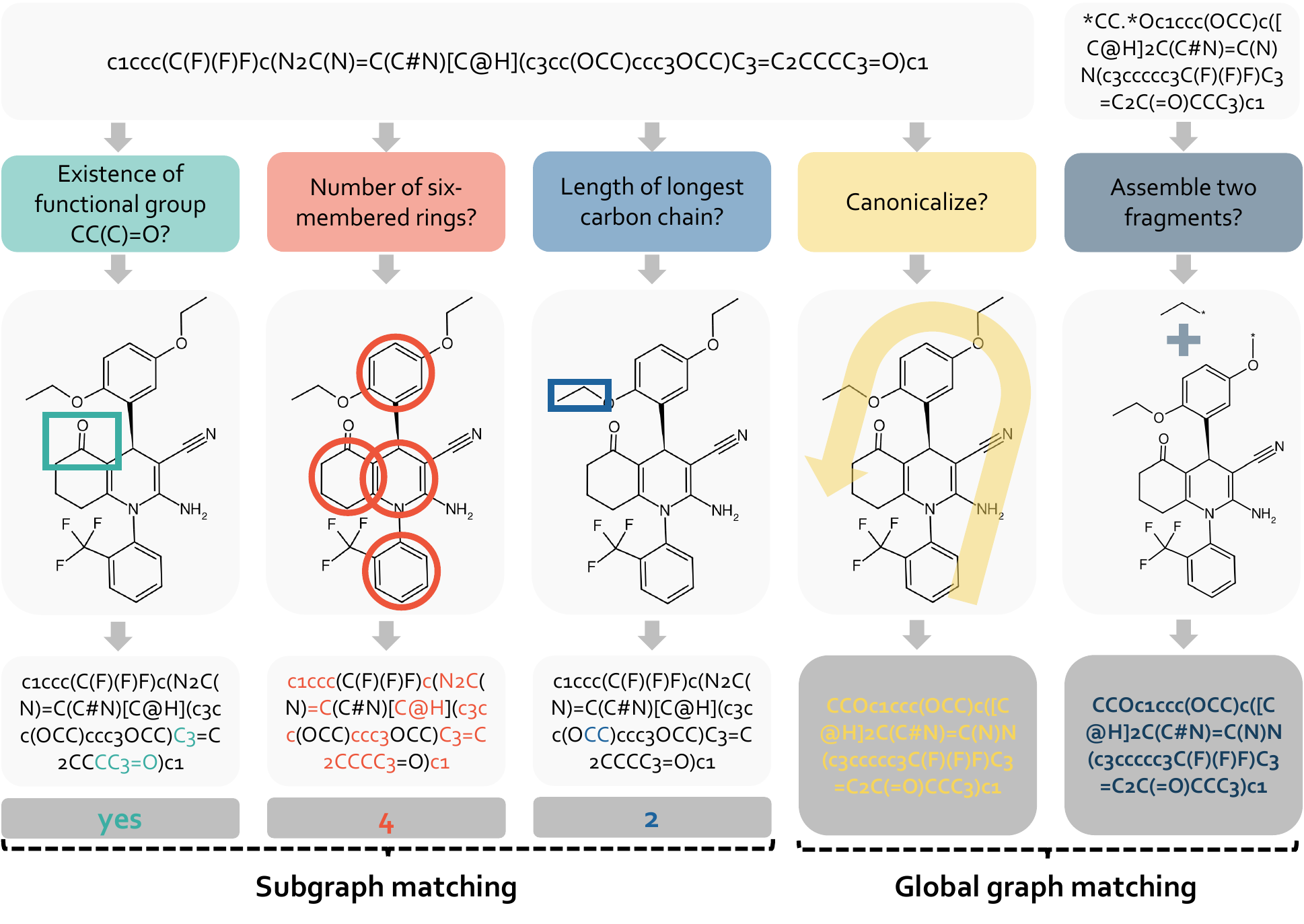}
    \vspace{-0.2in}
    \caption{\textbf{Illustration of SMILES parsing tasks.}}
    \label{fig:2_smiles_parsing_task}
    \vspace{-0.1in}
    \end{subfigure}
    \centering
    \begin{subfigure}[t]{0.34\textwidth}
        \includegraphics[width=\linewidth]{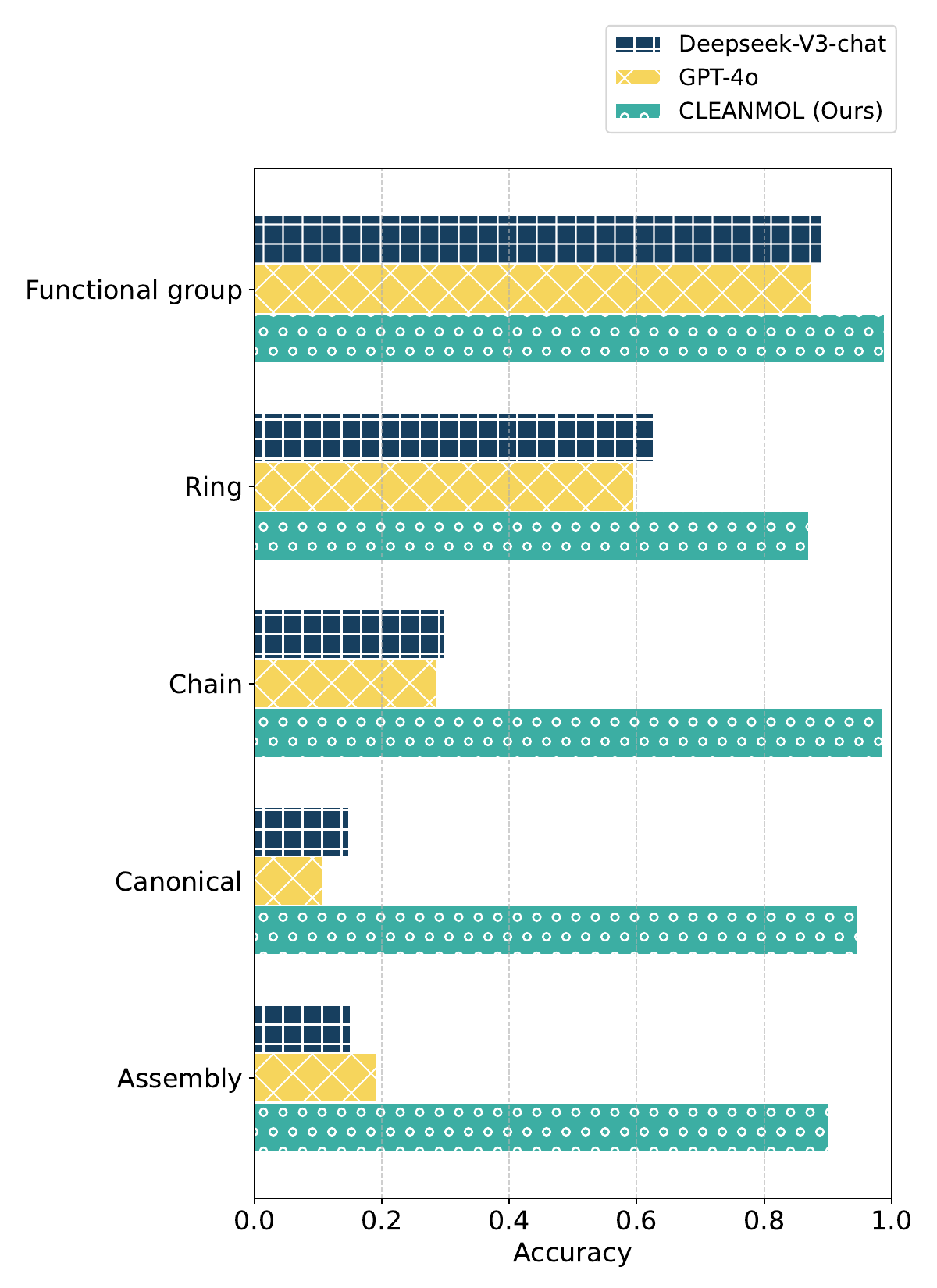}
        \vspace{-0.2in}
        \captionsetup{justification=centering}
        \caption{\textbf{Failure of LLMs on SMILES parsing.}}\label{fig:2_failure}
        \vspace{-0.1in}
    \end{subfigure}
    
    \caption{\textbf{Overview of SMILES parsing}. (\subref{fig:2_smiles_parsing_task}) Each column visualizes one of the five SMILES parsing tasks: \textcolor{mymint}{functional group matching}, \textcolor{myorange}{ring counting}, \textcolor{myblue}{carbon chain length measurement}, \textcolor{myyellow}{SMILES canonicalization}, and \textcolor{mynavy}{fragment assembly}. The highlighted tokens in the SMILES correspond to the substructures involved in each task. (\subref{fig:2_failure}) Recent LLMs fail for SMILES parsing while the model trained with our \Algname shows improvement.}\label{fig:1_parsing}
    \vspace{-0.2in}
\end{figure*}


%% file: 2.motivation_bottleneck.tex
\section{SMILES parsing task}

In this section, we introduce five SMILES parsing tasks designed to enhance the mapping between molecular SMILES strings and their corresponding graph structures. We then highlight two key bottlenecks in applying LLMs to molecular tasks: (1)~the inability of models to extract structural information from SMILES strings and (2) the lack of high-quality, scalable molecular datasets. To address the first bottleneck, we show that even advanced LLMs such as GPT-4o~\citep{openai2024gpt4technicalreport} and DeepSeek-V3~\citep{liu2024deepseek} fail to perform well on simple SMILES parsing tasks, revealing the need for explicit structure-aware supervision. To address the second bottleneck, we explain the limitation of open-source molecular datasets, motivating the need for scalable molecular datasets that can be generated without costly experiments.

\input{figure/fig_2_edge_case}

\subsection{SMILES parsing task description}\label{subsec:2_smiles_parsing}

We define SMILES parsing as a suite of deterministic, scalable, and structure-focused tasks designed to map molecular strings to their corresponding molecular graphs. The tasks fall into two categories—\textit{subgraph matching} and \textit{global graph matching}—as illustrated in \cref{fig:2_smiles_parsing_task}. Importantly, all annotations can be generated automatically using open-source chemical tools such as RDKit~\citep{greg2024rdkit} without any experiment, making the tasks highly scalable. We provide more details in \cref{appx:smiles_parsing}.

\begin{itemize}
    \item \textbf{Subgraph matching.} This category includes \textit{functional group matching}, \textit{ring counting}, and \textit{carbon chain length measurement}. Functional group matching determines the presence of a specified functional group. Ring counting identifies the number of rings with specific sizes (e.g., five- or six-membered), and chain length measurement evaluates the length of the longest carbon chain excluding rings. These tasks focus on local subgraphs such as structural motifs, branching, and ring patterns.

    \item \textbf{Global graph matching.} This category consists of \textit{SMILES canonicalization} and \textit{fragment assembly}. Canonicalization involves converting arbitrarily ordered SMILES into a canonical form, which encourages structural invariance to syntactic permutation. Fragment assembly requires the model to combine two SMILES fragments into a single valid molecule, testing its ability to reorganize the global structure from disjoint components.
\end{itemize}

\subsection{Failure of existing LLMs}\label{subsec:2_failure_smiles_parsing}

Although SMILES parsing appears simple from a structural point of view, it poses significant challenges for existing LLMs. Complex cases involving nested rings or hierarchical branching often disrupt token-level patterns, making it difficult for models to resolve SMILES parsing accurately. In detail, as shown in \cref{fig:2_edge_case}, many structural features are represented non-contiguously in SMILES, further complicating the parsing process. Our motivation closely aligns with that of \citet{jang2024chain}.\footnote{Unlike \citet{jang2024chain}, which fine-tunes models directly on structural information and downstream tasks, we pre-train LLMs on SMILES parsing objectives and subsequently fine-tune them for downstream tasks.}

\input{figure/fig_2_smiles_parsing}

We observe that even state-of-the-art general-purpose LLMs, including GPT-4o \citep{openai2024gpt4technicalreport} and DeepSeek-V3-Chat \citep{liu2024deepseek}, struggle with SMILES parsing, achieving no more than 60\% accuracy across five tasks except for the binary classification (functional group matching), as described in \cref{fig:2_failure} and detailed in \cref{subsec:4_1_smiles_parsing}. This failure is notable given the strong performance of these models in other domains such as mathematics and code. The inability of these models to handle even basic molecular parsing tasks underscores a critical gap in their structural understanding. It motivates the need for explicit pretraining strategies tailored to molecules.

\subsection{Costly high-quality data acquirement }\label{subsec:2_data_acquirement}

A second challenge lies in acquiring sufficient high-quality training data for molecules. In contrast to textual and visual domains, which benefit from large-scale web scraping \citep{deng2009imagenet, raffel2020exploring, lozhkov2024starcoder}, chemical datasets often rely on costly and labor-intensive wet lab experiments or computational simulations. While resources such as the USPTO series~\citep{wei2010novel, lu2022unified} and MoleculeNet~\citep{wu2018moleculenet} exist, expanding them is expensive and labor-intensive. This highlights the need for scalable alternatives—datasets that can be automatically generated with minimal cost while preserving domain relevance.

%% file: figure/fig_2_edge_case.tex
\begin{figure}
    \centering
    \includegraphics[width=\linewidth]{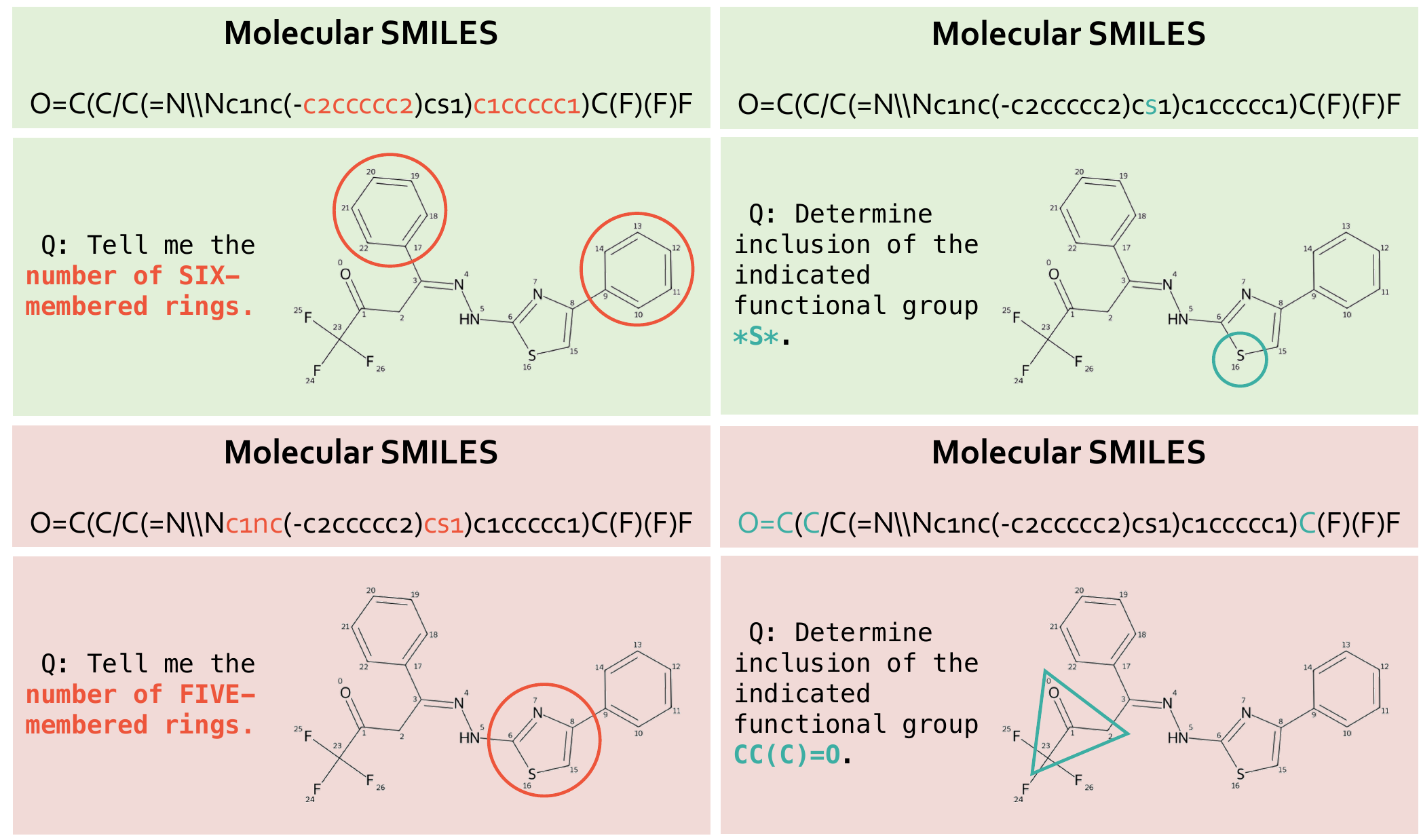}
    \vspace{-0.3in}
    \caption{\textbf{Complex cases in SMILES parsing.} The top green panels represent relatively simple cases, while the bottom red panels illustrate more complex examples with non-continuous substructures in SMILES. \textcolor{myorange}{Orange} and \textcolor{mymint}{teal} highlights correspond to tasks involving ring counting and functional group matching, respectively.}
    \label{fig:2_edge_case}
    \vspace{-0.2in}
\end{figure}

%% file: figure/fig_2_smiles_parsing.tex
\begin{figure*}
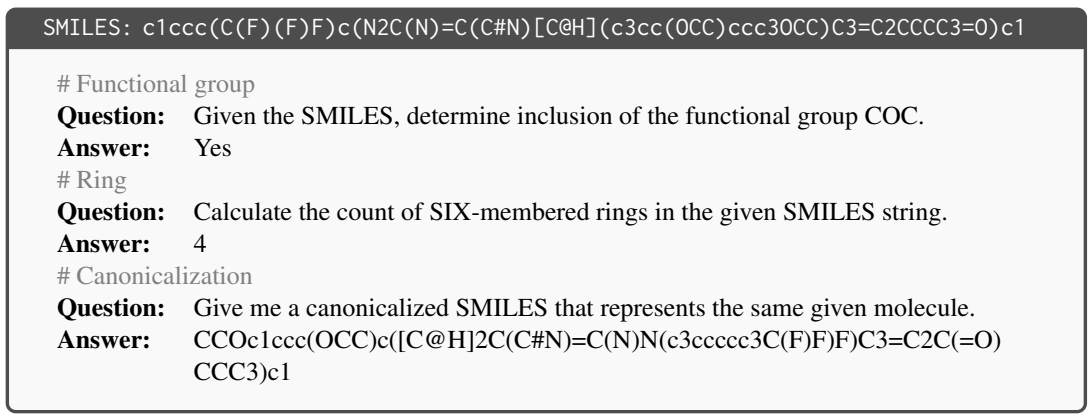

  \centering
  \resizebox{0.9\linewidth}{!}{
  \begin{tcolorbox}[
    colback=gray!5!white, 
    colframe=gray!60!black,
    title={
      \parbox[t]{\dimexpr\linewidth-4mm\relax}{%
        \ttfamily
        SMILES: c1ccc(C(F)(F)F)c(N2C(N)=C(C\#N)[C@H](c3cc(OCC)ccc3OCC)C3=C2CCCC3=O)c1

      }
    }
  ]
  \begin{tabularx}{\textwidth}{>{\bfseries}lX}
  \multicolumn{2}{l}{\textcolor{gray}{\# Functional group}} \\
    Question:& Given the SMILES, determine inclusion of the functional group COC.  \\
    Answer:& Yes \\
    \multicolumn{2}{l}{\textcolor{gray}{\# Ring}} \\
    Question:& Calculate the count of SIX-membered rings in the given SMILES string. \\
    Answer: & 4 \\
    \multicolumn{2}{l}{\textcolor{gray}{\# Canonicalization}} \\
    Question: & Give me a canonicalized SMILES that represents the same given molecule.\\
    Answer:  & CCOc1ccc(OCC)c([C@H]2C(C\#N)=C(N)N(c3ccccc3C(F)F)F)C3=C2C(=O) \\
    & CCC3)c1 \\

  \end{tabularx}
  \end{tcolorbox}
  }
  \vspace{-0.1in}
    \caption{\textbf{Examples of \Algname\ dataset.} }\label{fig:2_smiles_parsing}
    \vspace{-0.2in}
\end{figure*}

%% file: 3.method.tex
\input{figure/figure_3_pruning}

\section{Training framework of \Algname{}}\label{sec: method}

In this section, we present our framework to improve the molecular understanding of LLMs using a new dataset, coined \Algname.\footnote{Our framework and dataset are both termed \Algname.} Our scheme consists of (1) data preparation and (2) a two-stage training procedure. In the data preparation step, we prepare the \Algname dataset with deterministic and scalable SMILES parsing tasks. Next, in the training step, we pre-train LLMs with the \Algname dataset, followed by fine-tuning downstream applications. To improve the pre-training, we also introduce a task-adaptive data pruning and curriculum learning strategy based on task-specific difficulty measures.

\subsection{\Algname data preparation}\label{subsec: data_prepare}

First, we introduce the \Algname dataset based on the SMILES parsing tasks proposed in \cref{subsec:2_smiles_parsing}. There exist two key advantages of our proposed tasks: determinism and scalability. 

In detail, on the one hand, in terms of determinism, our tasks are designed to have a unique and clearly defined answer (i.e., number or canonicalized SMILES) unlike previous pre-training objectives such as masking and translation as detailed in \cref{sec: related}. This ensures unambiguous supervision during training and facilitates reliable learning.

 On the other hand, regarding scalability, as the proposed tasks apply to any valid molecules without any experimental data, they can be expanded to a vast set of molecules. In detail, all annotations can be automatically generated using open-source cheminformatics tools such as RDKit~\citep{greg2024rdkit}, making the dataset extensible to virtually unlimited molecular corpora. We provide the simplified example instructions of SMILES parsing tasks in \cref{fig:2_smiles_parsing} and more examples including detailed instruction formats in \cref{appx:smiles_parsing}.

\subsection{Training with \Algname}\label{subsec: training}

\input{table/tab_3_difficulty}

\input{table/tab_parsing}

Once the \Algname dataset is prepared, we adopt a task-specific \textbf{data pruning} and \textbf{curriculum learning} inspired by recent work on high-quality LLM data curation~\citep{gunasekar2023textbooks, marion2023less, ankner2024perplexed} to further enhance pre-training with \Algname{}. As illustrated in \cref{fig:3_data_pruning}, our approach involves: (1) subsampling sufficiently informative molecules, and (2) constructing a curriculum by ranking these examples from simple to complex using task-specific difficulty measures.

The difficulty measures are defined for each parsing task as summarized in \cref{tab:difficulty}. For instance, in the chain length measurement task, molecules with extensive branches often lead to SMILES where relevant subgraph atoms appear far apart in the string, increasing parsing difficulty. By excluding extremely easy or hard molecules (i.e., subsample molecules with mid-level difficulties) and organizing the training data from simple to complex, our approach aligns with curriculum learning principles \citep{bengio2009curriculum} and leads to improved performance, as validated in \cref{subsec: pruning}.

Next, we adopt a two-stage training pipeline to effectively integrate SMILES parsing into LLM. In the first stage, we perform pre-training on the pruned \Algname dataset using supervised fine-tuning. This allows the model to acquire core structural understanding and compositional knowledge of molecular graphs. In the second stage, we further fine-tune this trained model on downstream molecular tasks. By initializing with a model that has already learned to parse molecular structures, downstream adaptation becomes more accurate.

%% file: figure/figure_3_pruning.tex
\begin{figure}
    \centering
    \includegraphics[width=\linewidth]{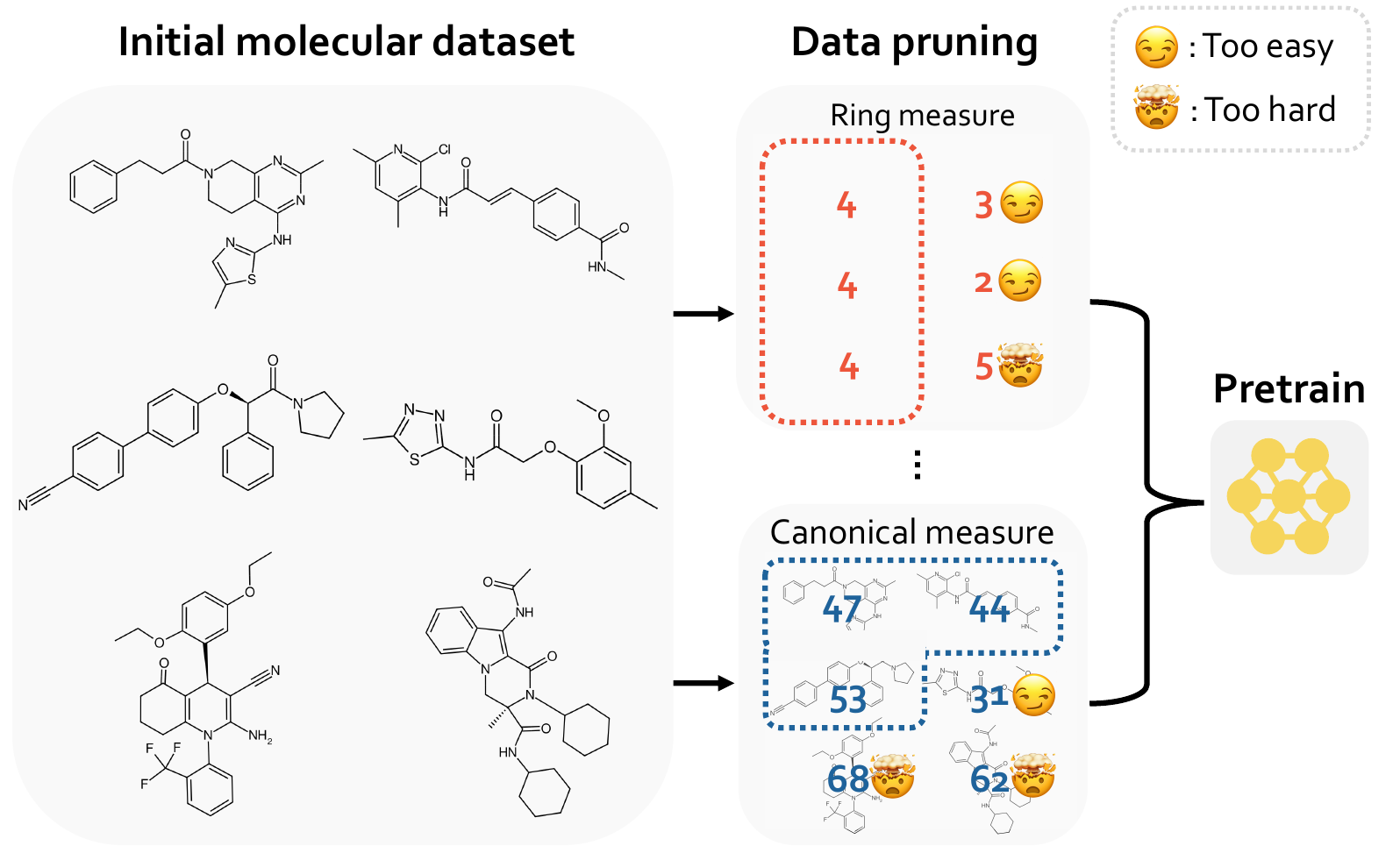}
    \vspace{-0.1in}
    \caption{\textbf{Overview of molecular data pruning and ranking.} Each number represents the task-specific difficulty score assigned to a molecule, as defined in \cref{tab:difficulty}. For each parsing task, molecules are ranked based on these scores and we select the mid-difficulty samples.}
    
    \label{fig:3_data_pruning}
    \vspace{-0.2in}
\end{figure}

%% file: table/tab_3_difficulty.tex
\begin{table}[t]
  \resizebox{\linewidth}{!}{%
    \begin{tabular}{ccccc}
      \toprule[1.25pt]
      \textbf{Functional group} & \textbf{Ring} & \textbf{Chain length} 
        & \textbf{SMILES} & \textbf{Fragment} \\
      \textbf{matching} & \textbf{counting} & \textbf{measurement} 
        & \textbf{canonicalization} & \textbf{assembly} \\
      \midrule
      \# of functional groups & \# of rings & \# of branches 
        & \multicolumn{2}{c}{SMILES length} \\
      \bottomrule[1.25pt]
    \end{tabular}%
  }
  \caption{\textbf{Definition of each task‐specific difficulty.}}\label{tab:difficulty}
  \vspace{-0.2in}
\end{table}

    
      
      

%% file: table/tab_parsing.tex
\newcolumntype{Y}{>{\centering\arraybackslash}X}

\begin{table*}[t]
  \centering
  \resizebox{0.7\linewidth}{!}{
  \begin{tabularx}{\textwidth}{@{} c c *{5}{Y} @{}}
    \toprule[1.25pt]
            &          
  & \multicolumn{3}{c}{\textbf{Subgraph}}               & \multicolumn{2}{c}{\textbf{Global graph}}   \\
    \cmidrule(lr){3-5}  \cmidrule(lr){6-7}
  \textbf{Task type}  & \textbf{Model} & \textbf{FG} 
      & \textbf{Ring} & \textbf{Chain}
      & \textbf{Canonical} & \textbf{Assembly}  \\
    \midrule[1.25pt]
    \multirow{3}{*}{5-shot}
      & Deepseek-V3-chat & 0.8912 & 0.6266 & 0.2976 & 0.1484 & 0.1512  \\
      & GPT-4o  & 0.8750 & 0.5955 & 0.2857 & 0.1078 & 0.1932  \\
      & Galactica-6.7B & 0.5000 & 0.0732 & 0.1511 & 0.0000 & 0.0046 \\
    \midrule
    \multirow{4}{*}{SFT}
      & Llama3.1-8B (Single) & 0.9414 & 0.8612 & 0.9859 & 0.9356 & 0.8858  \\
      & Llama3.1-8B (Multi)  & \cll{0.9891} & \cll{0.8707} & 0.9851 & \cll{\textbf{0.9463}} & \cll{\textbf{0.9010}} \\
      \cmidrule(lr){2-7}
      & Qwen2.5-7B (Single)  & 0.9891 & 0.8674 & \textbf{0.9907} & 0.7593 & 0.3371  \\
      & Qwen2.5-7B (Multi)   &  \cll{\textbf{0.9901}} & \cll{\textbf{0.8750}} & 0.9902 & \cll{0.9262} & \cll{0.8835} \\
    \bottomrule[1.25pt]
  \end{tabularx}
  }
  \vspace{-0.1in}
    \caption{\textbf{SMILES parsing performance.} FG stands for the functional group. \colorbox{blue!4}{Background} indicates the improvement of multi-task learning compared to the single-task learning and the best results are highlighted in \textbf{bold}.}\label{tab:smi_parsing}
    \vspace{-0.2in}
\end{table*}

%% file: 4-1.experiments_fundamental.tex
\section{Experiments: SMILES parsing tasks}\label{sec: exp_smiles_parsing}

In this section, we evaluate the effectiveness of our proposed SMILES parsing task as a pre-training signal for LLMs. The parsing task is formally defined in \cref{subsec:2_smiles_parsing}. We demonstrate that recent LLMs, while not inherently proficient in SMILES parsing, can acquire this capability through targeted training. We provide all experimental settings including prompts, hyperparameters, and computational resources in \cref{appx: exp}.

\subsection{LLMs can learn SMILES parsing}\label{subsec:4_1_smiles_parsing}

\input{table/tab_sort}

As described in \cref{subsec:2_failure_smiles_parsing}, SMILES parsing poses a significant challenge for general-purpose LLMs, despite its foundational importance for molecular understanding. Our experiments reveal that LLMs lack the inductive bias to naturally understand the molecular structure encoded in SMILES strings. However, we show that through supervised fine-tuning (SFT), LLMs can learn to accurately parse and interpret SMILES representations. 

\paragraph{Dataset.} We construct a \Algname{} benchmark consisting of 50K molecules per SMILES parsing task, totaling 250K examples across five tasks. The molecules are subsampled from the ZINC250k \citep{irwin2012zinc} training dataset using our proposed molecular data pruning strategy described in \cref{subsec: training}, which excludes extremely easy or hard molecules to enhance the molecular pre-training. Additionally, for the test dataset, we randomly selected 10K molecules from the ZINC250K test split and fixed this subset across all experiments.

\paragraph{Baselines.} We evaluate the parsing capabilities of four general-purpose LLMs—Deepseek-V3-Chat~\citep{liu2024deepseek}, GPT-4o~\citep{openai2024gpt4technicalreport}, LLaMA3.1-8B-Instruct~\citep{grattafiori2024llama}, and Qwen2.5-7B-Instruct~\citep{yang2024qwen2}—and one chemistry-specific LLM, Galactica-6.7B~\citep{taylor2022galactica}. To assess the basic molecular understanding of general-purpose LLMs, we apply 5-shot prompting to Deepseek and GPT-4o, which are not publicly trainable and thus cannot be fine-tuned. Similarly, we apply 5-shot prompting to Galactica, a chemistry-specific LLM pre-trained on molecular corpora, to evaluate its zero-shot capabilities without further supervision. In contrast, for LLaMA and Qwen, which are open-weight general-purpose LLMs, we perform supervised fine-tuning using our SMILES parsing dataset to examine whether explicit structure-aware training can bridge the gap in molecular comprehension. Notably, we explore two experimental settings: \textit{single-task}, where a separate model is trained for each parsing task, and \textit{multi-task}, where a single model is jointly trained on all five tasks.

\paragraph{Metrics.} We evaluate performance using accuracy, as SMILES parsing tasks are deterministic and each input has a well-defined answer.

\paragraph{Results.} The results are presented in \cref{tab:smi_parsing}. We observe that recent general-purpose LLMs (GPT-4o and Deepseek) and even a chemical LLM (Galactica) perform poorly on SMILES parsing, revealing their limited molecular comprehension. This validates that \textit{the primary bottleneck in applying LLMs to molecular domains lies not in the absence of chemical knowledge, but in the lack of basic molecular structural understanding—specifically, the ability to parse and interpret SMILES strings}. In contrast, fine-tuned LLaMA and Qwen models show substantial improvements, demonstrating that SMILES parsing can be effectively learned through training. Moreover, all tasks—except for chain length measurement—achieved higher accuracy in the multi-task setting, suggesting that transferable structural understanding across tasks contributes to improved performance.

\subsection{Effect of molecular data pruning}\label{subsec: pruning}

We further investigate the impact of our molecular data pruning strategy on parsing performance. As detailed in \cref{subsec: training}, this technique aims to curate a training set that maximizes informativeness. The results, shown in \cref{tab:sort}, demonstrate that our pruning method improves performance, suggesting that data quality plays a critical role in teaching LLMs the implicit grammar of SMILES.

\input{figure/fig_4_ablation}

\subsection{Ablation study}

Here, we conduct an ablation study to validate the impact of the increase in dataset size in our proposed \Algname dataset. In detail, we evaluate the accuracy of the SMILES parsing task for 10K, 20K, and 50K data settings per task in the same setting in \cref{subsec:4_1_smiles_parsing}. We provide the results in \cref{fig:4_ablation}. Here, we observed that increasing the dataset size consistently improves SMILES parsing performance, with particularly dramatic gains in the ring counting and fragment assembly tasks. This validates the expandability of our framework.

%% file: table/tab_sort.tex
\newcolumntype{Y}{>{\centering\arraybackslash}X}

\begin{table*}[t]
  \centering
  \resizebox{0.7\linewidth}{!}{%
    \begin{tabularx}{\linewidth}{@{} c *{5}{Y} c}
      \toprule[1.25pt]
      & \multicolumn{3}{c}{\textbf{Subgraph}}
      & \multicolumn{2}{c}{\textbf{Global graph}}
      &  \\
    \cmidrule(lr){2-4} \cmidrule(lr){5-6}
      \textbf{Pruning type}
        & \textbf{FG}
        & \textbf{Ring}
        & \textbf{Chain}
        & \textbf{Canonical}
        & \textbf{Assembly}
        & \textbf{Average} \\    
        \midrule[1.25pt]

      Random               & {0.9921} & 0.9212 & 0.9886 & 0.7845 & 0.7352 & 0.8843 \\
      Length               & 0.9910 & 0.8531 & 0.9785 & 0.8519 & 0.8044 & 0.8958 \\
    Molecular pruning (top)   & {0.9902} & 0.8123 & 0.9716
                           & {0.9446} & 0.7487 & 0.8934 \\
      Molecular pruning (bottom)& 0.9729 & 0.6995 & 0.9597
                           & 0.5514 & 0.5186 & 0.7404 \\
       \midrule
     Molecular pruning (middle, \textbf{ours})& 0.9901 & {0.8750} & {0.9902}
                           & 0.9262 & {0.8835} & \textbf{0.9330} \\
      
      \bottomrule[1.25pt]
    \end{tabularx}
  }
  \vspace{-0.1in}
  \caption{\textbf{Effect of molecular data pruning on Qwen2.5-7B-Instruct.} "Random" and "Length" refer to baselines using random sampling and SMILES length as proxies for difficulty. "Top," "middle," and "bottom" denote subsamples consisting of the most difficult, moderately difficult, and easiest molecules, respectively, based on task-specific difficulty heuristics. }
  \label{tab:sort}
  \vspace{-0.2in}
\end{table*}

%% file: figure/fig_4_ablation.tex
\begin{figure}
    \centering
    \includegraphics[width=0.9\linewidth]{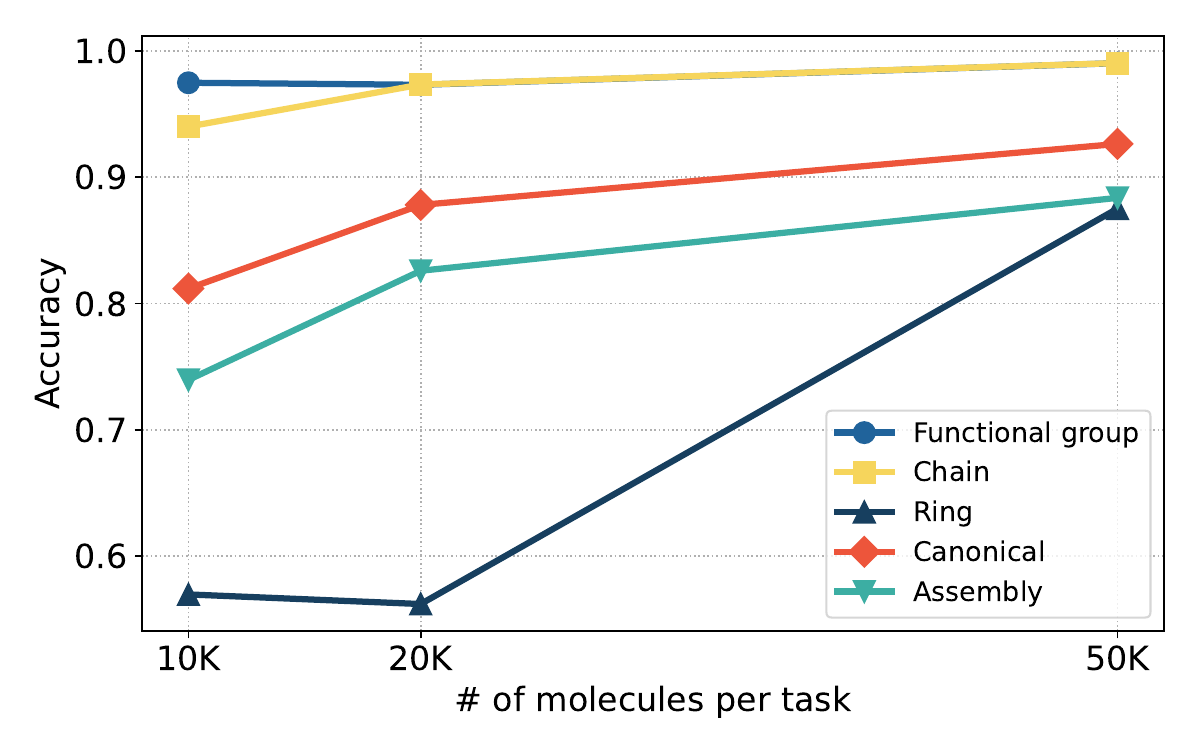}
    \vspace{-0.1in}
    \caption{\textbf{Data scale analysis for SMILES parsing.}}
    \label{fig:4_ablation}
    \vspace{-0.2in}
\end{figure}

%% file: 4.experiments.tex
\section{Experiments: Downstream tasks}\label{sec: downstream_task}

In this section, we evaluate the effect of pre-training LLMs on \Algname dataset across three molecular generation downstream applications. We provide the experimental settings in \cref{appx: exp} and additional experimental results in \cref{appx: sample}.

\input{table/tab_mol_gen}

Our results demonstrate that incorporating \Algname{} as a pre-training strategy consistently improves performance across diverse downstream molecular tasks. These findings provide strong empirical support for our central hypothesis: clean and structurally faithful SMILES parsing serves as an effective and transferable learning signal for LLMs. Notably, \Algname{} achieves state-of-the-art or competitive performance despite being pre-trained without any task-specific data, underscoring the strength and generality of our approach.

\subsection{Molecular generation}\label{subsec:5_molgen}
The molecular generation task aims to generate molecules given prompts, including retrosynthesis, reagent prediction, and forward reaction prediction.

\paragraph{Dataset.} We use the Mol-Instructions dataset ~\citep{fang2024molinstructions}, which covers three molecule generation tasks. Specifically,  retrosynthesis predicts the possible precursors that lead to a given target molecule. Next, the reagent prediction task requires the generation of suitable catalysts, solvents, or ancillary reagents for a given chemical reaction. Lastly, forward reaction prediction involves the generation of a plausible product from given reactants and reagents. We follow the data splits provided in Mol-Instructions.

\paragraph{Baselines.} We evaluate \Algname by integrating it with two base models: LLaMA-3.1-8B-Instruct \citep{grattafiori2024llama} and Qwen-2.5-7B-Instruct \citep{yang2024qwen2}, to test whether \Algname consistently improves performance. Notably, the vanilla base models are fine-tuned on each downstream task without pre-training. For an absolute performance comparison, we include three baselines: Text+Chem T5 \citep{christofidellis2023chemt5}, Mol-Instructions \citep{fang2024molinstructions} and InstructMol \citep{cao2023instructmol}. Additionally, we include a variant of Mol-Instructions denoted as Mol-Instructions (SFT), which is first instruction-tuned on the same dataset size as our \Algname dataset (250K) and then further fine-tuned on each downstream task. This ensures a fair comparison for both the model and the training data size.

\paragraph{Metrics.} We assess the performance by comparing the generated molecules with the ground truth based on eight metrics. These include SMILES string-based metrics (Exact match, BLEU~\citep{papineni2002bleu}, and Levenshtein distance \citep{miller2009levenshtein}), molecular fingerprint similarities (MACCS \citep{durant2002maccs}, RDK \citep{schneider2015rdk}, and Morgan \citep{rogers2010morgan}), distributional similarity via Fréchet ChemNet Distance (FCD) \citep{preuer2018fcd}, and the validity of generated molecules.

\paragraph{Results.} The results are summarized in \cref{tab: mol_gen}. Incorporating \Algname consistently improves performance across all backbones, demonstrating the effectiveness of SMILES parsing tasks in enhancing molecular language modeling. These improvements suggest that pre-training on clean and deterministic \Algname{} dataset facilitates the model’s structural understanding required for generation tasks. Notably, integrating \Algname into LLaMA3.1-8B-Instruct achieves state-of-the-art—or at least comparable—performance to Mol-Instructions (SFT), despite using no molecular generation data during pre-training.

\subsection{Ablation study}

\input{figure/fig_5_ablation}

Here, we evaluate the effect of \Algname dataset size on retrosynthesis performance using 10K, 20K, and 50K molecules per parsing task following the setup in \cref{subsec:5_molgen}. As described in \cref{fig:5_ablation}, the performance grows with data scale, demonstrating \Algname’s scalability. As SMILES parsing requires no costly experiment, this framework easily extends to large molecular corpora.

%% file: table/tab_mol_gen.tex
\begin{table*}[t]
  \vspace{-0.2in}
  \centering
  \resizebox{0.84\linewidth}{!}{
    \begin{tabular}{cccccccc}
    \toprule[1.25pt]
      Models & \textbf{Exact.} & \textbf{BLEU} & \textbf{Levenshtein} $\downarrow$ & \textbf{MACCS FTS} & \textbf{RDK FST} & \textbf{Morgan FTS} & \textbf{Validity} \\
\midrule[1.25pt]
    \rowcolor{whitegray} \multicolumn{8}{l}{\textit{Task 1: Retrosynthesis}} \\
    \midrule
    Text+Chem T5 & 0.141 & 0.765 & 24.04 & 0.685 & 0.765 & 0.585 & 0.698 \\
     Mol-Instructions (Lla.2) & 0.009  & 0.705 & 31.23 & 0.283 & 0.487 & 0.230 & - \\
     Mol-Instructions (Lla.3) & 0.333 & 0.842 & 17.64 & 0.704 & 0.815 & 0.646 & - \\
     Mol-Instructions (Lla.3.1)* & 0.255 & 0.890 & 17.76 & 0.813 & 0.690 & 0.644 & - \\
     InstructMol-GS & 0.407 & 0.941 & 13.97 & 0.753 & 0.852 & 0.714 & - \\
      \midrule
     Llama3.1-8B & 0.456 & 0.944 & 10.22 & 0.895 & 0.837 & 0.801 & 0.979 \\
     \quad + Mol-Instructions (SFT)* & \cll 0.541 & \cll 0.955 & \cll \phantom{0}\underline{8.25} & \cll \underline{0.915} & \cll 0.878 & \cll 0.843 & - \\
     \quad + \Algname & \cll{\textbf{0.581}} & \cll{\textbf{0.959}} & \phantom{0}\cll{\textbf{7.86}} & \cll{\textbf{0.923}} & \cll{\textbf{0.890}} & \cll{\textbf{0.856}} & \cll{\textbf{0.998}} \\
    \midrule
     Qwen2.5-7B & 0.460 & 0.946 & 10.11 & 0.897 & 0.849 & 0.809 & 0.910   \\
     \quad + \Algname & \cll{\underline{0.554}} & \cll{\underline{0.958}} & \phantom{0}\cll{8.26} & \cll{\underline{0.915}} & \cll{\underline{0.880}} & \cll{\underline{0.844}} & \cll{\underline{0.995}} \\
         \midrule[1.25pt]
    \rowcolor{whitegray} \multicolumn{8}{l}{\textit{Task 2: Reagent prediction}} \\
    \midrule
    Text+Chem T5 & 0.000 & 0.255 &49.32 & 0.039 & 0.186 & 0.052 & 0.313 \\
     Mol-Instructions (Lla.2) & 0.044 & 0.224 & 23.17 & 0.237 & 0.364 & 0.213 & - 
     \\
     Mol-Instructions (Lla.3) & 0.101 & 0.648 & 18.33 & 0.412 & 0.521 & 0.375 &  -  \\ 
     Mol-Instructions (Lla.3.1)* & 0.085 & 0.676 & 22.40 & 0.505 & 0.398 & 0.356 & - \\

    InsturctMol & 0.129 & 0.610 & 19.66 & 0.444 & \textbf{0.539} & 0.400 & - \\
     \midrule
     Llama3.1-8B & 0.124 & 0.625 & 17.31 & 0.538 & 0.433 & 0.398 & \textbf{0.999} \\
     \quad +      Mol-Instructions (SFT)* & \cll \underline{0.142} &\cll 0.678 &\cll 17.14 & \cll \underline{0.562} & \cll 0.467 & \cll \underline{0.430} & - \\
     \quad + \Algname & \cll{\textbf{0.147}} & \cll{\textbf{0.687}} & \cll{\underline{16.89}} & \cll{\textbf{0.564}} & \cll{\underline{0.472}} & \cll{\textbf{0.434}} & \cll{\textbf{0.999}} \\
    \midrule
     Qwen2.5-7B & 0.120 & 0.649 & 17.76 & 0.533 & 0.431 & 0.395 & - \\
     \quad + \Algname & \cll{0.128} & \cll{0.685} & \cll{\textbf{16.58}} & \cll{0.557} & \cll{0.455} &  \cll{0.415} & \underline{0.975} \\
     
     \midrule[1.25pt]
    \rowcolor{whitegray} \multicolumn{8}{l} 
    {\textit{Task 3: Forward reaction prediction}} \\
    \midrule
     Text+Chem T5 & 0.236 & 0.782 & 13.63 & 0.523 & 0.630 & 0.505 & 0.967 \\
    Mol-Instructions (Lla.2) & 0.045 & 0.654 & 27.26 & 0.313 & 0.509 & 0.262 & - \\
    Mol-Instructions (Lla.3) & 0.503 & 0.883 & 13.41 & 0.756 & 0.863 & 0.708 & -  \\
    Mol-Instructions (Lla.3.1)* & 0.402 & 0.907 & 13.11 & 0.848 & 0.718 & 0.679 & - \\
    InstructMol-GS & 0.536 & 0.967 & 10.85 & 0.776 & 0.878 & 0.741 & - \\
     \midrule
     Llama3.1-8B & 0.794 & 0.981 & \phantom{0}2.47 & 0.965 & 0.938 & 0.926 & \underline{0.988} \\
     \quad +  Mol-Instructions (SFT)* & \cll \underline{0.888} & \cll \textbf{0.990} & \cll \phantom{0}\textbf{1.33} & \cll \textbf{0.983} & \cll \textbf{0.967} & \cll \textbf{0.961} & - \\
     \quad + \Algname & \cll{\textbf{0.890}} & \cll{\textbf{0.990}} & \phantom{0}\cll{\underline{1.37}} & \cll{\underline{0.980}} & \cll{\underline{0.966}} & \cll{\underline{0.959}} & \cll{\textbf{0.996}} \\
    \midrule
     Qwen2.5-7B & 0.833 & 0.986 & \phantom{0}2.08 & 0.972 & 0.947 & 0.943 & 0.987 \\
     \quad + \Algname & \cll{0.874} &  \cll{0.989} &\phantom{0}\cll{1.56} & \cll{\underline{0.980}} & \cll{0.963} & \cll{0.956} & 0.959 \\

    \bottomrule[1.25pt]
  \end{tabular}
  }
  \vspace{-0.1in}
  \caption{\textbf{Molecular generation performance.} 
  \colorbox{blue!4}{Background} indicates the improvement compared to vanilla model. Asterisks (*) denote reproduced results and - in validity represents the SELFIES-based methods which guarantees the perfect validity. For each metric, the best and second-best result is highlighted with \textbf{bold} and \underline{underline}.} \label{tab: mol_gen}
 \vspace{-0.2in}
\end{table*}

%% file: figure/fig_5_ablation.tex
\begin{figure}
    \centering
    \includegraphics[width=0.9\linewidth]{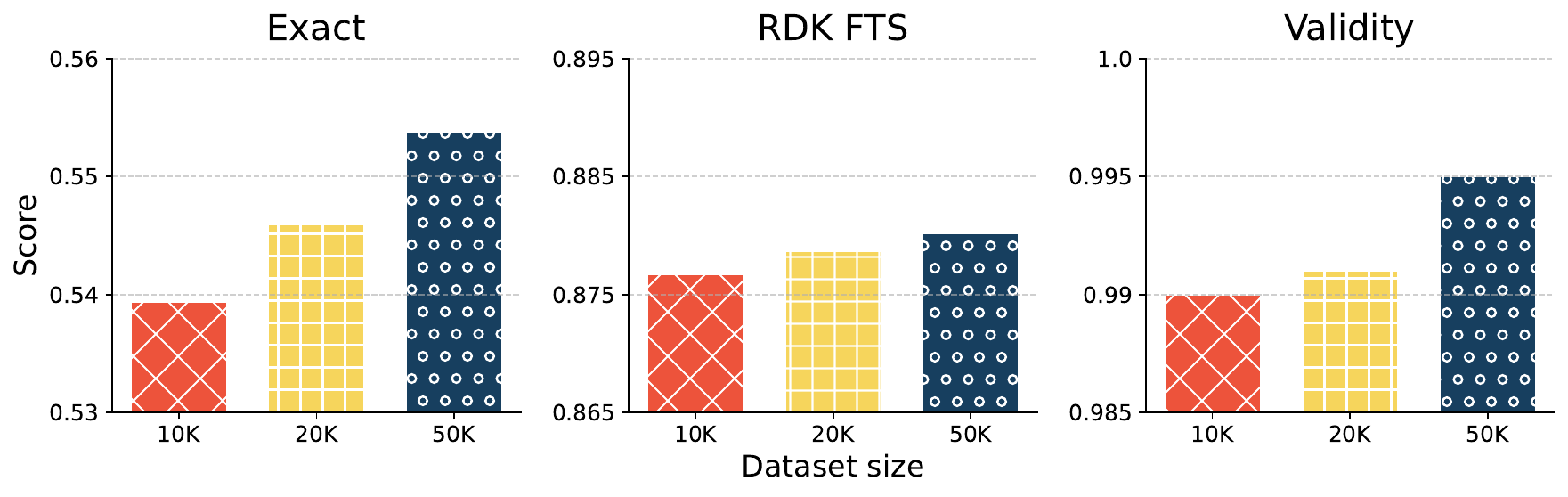}
    \vspace{-0.1in}
    \caption{\textbf{Data scale analysis for retrosynthesis.}}
    \label{fig:5_ablation}
    \vspace{-0.2in}
\end{figure}

%% file: 5.related.tex
\section{Related work}\label{sec: related}

\paragraph{LLMs for chemistry.} General-purpose LLMs often struggle with fundamental chemistry tasks, particularly those requiring molecular structure understanding \citep{white2023assessment, nascimento2023understandchemistry, guo2023can}. To address this gap, several studies have proposed chemically specialized LLMs. Some approaches pre-train LLMs on molecular and biomedical corpora to inject domain-specific knowledge \citep{edwards2022molt5, christofidellis23unifying, liu-etal-2023-molxpt, pei-etal-2023-biot5}. Others explore instruction tuning on curated molecular tasks \citep{fang2024molinstructions, cao2023instructmol}, or leverage retrieval-augmented prompting to improve few-shot performance \citep{Li2024molregpt}. While these methods aim to inject domain knowledge, they often neglect the need for grounding models in basic molecular understanding. In contrast, we emphasize clean and deterministic structural supervision through well-defined SMILES parsing tasks, which can complement existing methods and integrate with instruction tuning or domain adaptation.

\paragraph{Pre-training of LLMs for chemistry.} Effective pre-training tasks should be well-structured and sufficiently simple to support generalizable learning. In chemistry, many works adopt NLP-inspired objectives such as masked language modeling (MLM)~\citep{devlin2019bert} and sequence-to-sequence translation~\citep{raffel2020t5}, applied to SMILES~\citep{weininger1988smiles} or SELFIES~\citep{krenn2020self}. \citet{edwards2022molt5} used separate MLM pretraining on molecular and textual data, while later studies~\citep{pei-etal-2023-biot5, christofidellis23unifying} combined MLM with molecule–text translation. \citet{liu-etal-2023-molxpt} embedded SMILES in natural language prompts, and other works incorporated 2D or 3D geometry~\citep{li2023knowledge, ji2024exploring, zhou2023unimol}.

Despite these advancements, most strategies introduce unambiguous supervision signals due to the non-determinism of molecular representations. For example, in masked SMILES prediction, multiple chemically valid tokens can fill the same masked position, leading to a noisy training signal. This undermines training effectiveness and limits the model’s ability to learn robust understanding. To address this issue, we provide clean and deterministic SMILES parsing tasks as pre-training tasks.

\paragraph{Data pruning in LLMs.} Data pruning refers to selecting an informative subset of training data, which is crucial for reliable LLM training~\citep{gunasekar2023textbooks}. Most data pruning methods rely on rule-based filters~\citep{wenzek-etal-2020-ccnet, raffel2020exploring}, perplexity scores~\citep{marion2023less, ankner2024perplexed}, or LLM embeddings~\citep{tirumala2023d4}. However, these metrics are ill-defined for molecular strings, where perplexity and embeddings do not reflect the structural information of the corresponding molecules. To address this, we introduce task-specific difficulty measures and data pruning strategies for molecules.

%% file: 5.conclusion.tex
\section{Conclusion}

In this paper, we revisit the key limitation in applying LLMs to chemistry: the inability to interpret the structures encoded in SMILES. To address this, we propose \Algname, a framework that introduces deterministic and scalable SMILES parsing tasks to provide unambiguous structural supervision. Our experiments show that \Algname significantly enhances molecular structural understanding and improves performance across multiple downstream tasks. These results highlight the value of incorporating clean and structure-aware objectives into LLMs to support more robust applications.

%% file: 8.appendix.tex
\appendix
\begin{center}{\bf {\LARGE Appendix}}\end{center}

\paragraph{Organization} The appendix is organized as follows: We first describe the details of SMILES parsing tasks in \cref{appx:smiles_parsing}. Next, we present the experimental details such as hyperparameters and computational resources in \cref{appx: exp}. Then we provide the additional experimental results including the generated samples and additional ablation studies in \cref{appx: sample}. Lastly, we present the usage of AI assistants and scientific artifacts in \cref{appx: ai_assistant} and \cref{appx: artifact}, respectively.

\section{Detailed description of SMILES parsing tasks}\label{appx:smiles_parsing}

\subsection{Subgraph matching}

This category includes \textit{functional group matching}, \textit{ring counting}, and \textit{carbon chain length measurement}. These tasks are designed to focus on local substructures within the molecular graph, such as common functional motifs, ring systems, and chain connectivity. Each task formulation is deterministic and lends itself to clear evaluation.

\paragraph{Functional group matching.} Functional group matching evaluates whether a specified functional group is present in a given molecule. To ensure determinism, we cast this task as a binary classification problem: the model must predict “yes” or “no” based on the presence of the target group. An example of the instruction format is shown in \cref{fig:appx_2_1_fg}.

\input{figure_appx/fig_appx_2_1_fg}

\paragraph{Ring counting.} Ring counting asks the model to determine the number of rings of a specific size (e.g., five- or six-membered) in the molecule. This task tests the model’s ability to track topological cycles through non-contiguous token spans in SMILES. The instruction format is illustrated in \cref{fig:appx_2_2_ring}.

\input{figure_appx/fig_appx_2_2_ring}

\paragraph{Chain length measurement.} This task requires the model to identify the length of the longest acyclic carbon chain in the molecule, excluding atoms that are part of rings. It challenges the model to distinguish between linear and branched motifs and to reason about connectivity beyond localized tokens. Such chains often span long syntactic distances in SMILES, making the task non-trivial. The instruction format is shown in \cref{fig:appx_2_3_chain}.

\input{figure_appx/fig_appx_2_3_chain}

\subsection{Global graph matching} This category includes tasks that operate on a global level: \textit{SMILES canonicalization} and \textit{fragment assembly}. Unlike subgraph matching, these tasks require full-graph interpretation, where success depends on integrating information across the entire molecular structure. 

This category consists of \textit{SMILES canonicalization} and \textit{fragment assembly}.

\paragraph{SMILES canonicalization.} Canonicalization involves transforming a randomly ordered SMILES string into its canonical form following the canonicalization rules \citep{weininger1989smiles}. In detail, these rules typically involve assigning a unique ranking to atoms based on graph invariants (e.g., atomic number, connectivity, bond types), selecting the lexicographically smallest traversal path, and applying consistent numbering for ring closures. This task encourages the model to learn structural invariance under permutation and reinforces a graph-level understanding of molecular identity.  The task format is provided in \cref{fig:appx_2_4_cano}.

\input{figure_appx/fig_appx_2_4_cano}

\paragraph{Fragment assembly.} Fragment assembly evaluates whether the model can reconstruct a full molecule from two disconnected SMILES fragments. This task tests global molecular coherence and the model’s ability to resolve attachment points into a chemically valid structure.  The instruction format of the instruction is shown in \cref{fig:appx_2_5_assemble}.

\input{figure_appx/fig_appx_2_5_assemble}

\section{Experimental details}\label{appx: exp}

In this section, we provide the details of the experiments. All experimental code related to this paper is available at \url{https://anonymous.4open.science/r/CLEANMOL} and our experiments are based on a single run. We use NVIDIA A100-80GB GPUs. We also apply low rank adaptation \citep{hu2022lora} and report results from a single run. Our implementations are based on the \texttt{transformers} library \citep{wolf-etal-2020-transformers}, the \texttt{trl} library \citep{vonwerra2022trl}, the \texttt{accelerate} library \citep{accelerate}, and \texttt{unsloth} library \citep{unsloth}. Additionally, we used the packages including rouge-score==0.1.2 and nltk==3.8.1.

\subsection{SMILES parsing}\label{appx: exp_anal}

Here, we describe the detailed settings for the SMILES parsing experiments in \cref{sec: exp_smiles_parsing}, including the pre-trainig step with SMILES parsing tasks. 

\paragraph{Hyperparameters.} The hyperparameters for all the models are provided in \cref{tab:appx_1_hyper_parsing}. We share the same hyperparameter for all the SMILES parsing tasks and base models. Notably, the model trained with SMILES parsing tasks is used as the pre-trained model for downstream tasks in \cref{sec: downstream_task}.

\input{table_appx/tab_appx_1_hyper_parsing}

\subsection{Downstream tasks}

Here, we describe the detailed settings for the downstream task experiments in \cref{sec: downstream_task}.

\paragraph{Hyperparameters.} The hyperparameters for all the models are provided in \cref{tab:appx_1_hyper_parsing}. We share the same hyperparameter for all downstream tasks and base models. Notably, for the reproduced Mol-instructions \citep{fang2024molinstructions} models, we follow the hyperparameters given in the original paper.

\input{table_appx/tab_appx_1_hyper_downstream}



\section{Additional experimental results}\label{appx: sample}
In this section, we provide additional experimental results including several concrete examples of generated samples.

\subsection{Molecular property prediction}

The molecular property classification task aims to predict binary labels for intrinsic physical or chemical properties, such as blood-brain barrier permeability or toxicity.

\paragraph{Dataset.} We use the MoleculeNet~\citep{wu2018moleculenet} dataset, focusing on three binary classification tasks: BACE, HIV, and Clintox. The BACE task predicts whether a molecule can inhibit human $\beta$-secretase 1 (BACE-1). The HIV task involves predicting the ability of compounds to inhibit HIV replication. The Clintox task assesses whether a compound is likely to fail clinical trials due to toxicity. We follow the splits provided in MoleculeNet.

\paragraph{Baselines.} We evaluate \Algname by integrating it with two base models: LLaMA-3.1-8B-Instruct \citep{grattafiori2024llama} and Qwen-2.5-7B-Instruct \citep{yang2024qwen2}. For an absolute performance comparison, we include additional baselines: MolCA \citep{liu-etal-2023-molca}, LlasMol \citep{yu2024llasmol} and InstructMol \citep{cao2023instructmol}. 

\paragraph{Metrics.} We evaluate the performance using accuracy, which denotes the overall proportion of correct predictions.

\input{table/tab_molecular_classification}

\paragraph{Results.} We report the results in \cref{tab:mol_classification}. We observe that models pre-trained with \Algname achieve consistent gains, confirming that the structural alignment learned during SMILES parsing transfers effectively to property classification tasks. 

\subsection{Molecular property regression}

The molecular property regression task focuses on predicting continuous-valued molecular properties.

\paragraph{Dataset.} We again use the Mol-Instructions \citep{fang2024molinstructions} dataset. We target quantum mechanics properties: HOMO energy, LUMO energy, and the energy gap (HOMO–LUMO difference). We also follow the same split.

\paragraph{Baselines} We evaluate \Algname by integrating it with two base models: LLaMA-3.1-8B-Instruct \citep{grattafiori2024llama} and Qwen-2.5-7B-Instruct \citep{yang2024qwen2}. For an absolute performance comparison, we include additional baselines: Alpaca \citep{alpaca-lora}, Baize \citep{xu-etal-2023-baize}, Vicuna \citep{chiang2023vicuna}, Galactica \citep{taylor2022galactica}, and Mol-Instructions \citep{fang2024molinstructions}. Here, the Mol-Instructions (SFT) follows the same training strategy described in \cref{subsec:5_molgen}.

\paragraph{Metrics.} We use mean absolute error (MAE) to evaluate prediction accuracy.

\input{table/tab_mol_reg}

\paragraph{Results.} We report the results in \cref{tab:mol_reg}. The results indicate that models pre-trained on SMILES parsing consistently outperform baselines, demonstrating that structural information learned via parsing enhances quantitative property prediction.

\section{Usage of AI assistants}\label{appx: ai_assistant}

In preparing this work, we used AI-based writing assistants to improve sentence structure, correct grammatical errors, and enhance overall readability. These tools were employed solely for language refinement and did not contribute to the development of technical content, research methodology, or experimental analysis. All scientific ideas, results, and conclusions presented in the paper were conceived and authored entirely by the researchers. The use of AI assistance was restricted to editorial purposes and did not affect the originality or intellectual contributions of the work.

\section{Scientific Artifacts}\label{appx: artifact}

\paragraph{The License for artifacts.} All datasets and software tools used in this study comply with their respective licenses. Specifically, we utilized publicly available datasets such as ZINC250K~\citep{irwin2012zinc} and Mol-Instructions~\citep{fang2024molinstructions} in accordance with their usage terms. External tools such as RDKit were employed under their permissive open-source license. To support transparency and reproducibility, we release our trained models and source code at \url{https://anonymous.4open.science/r/CLEANMOL} under an appropriate open-source license.

\paragraph{Artifact use consistency with intended use.} All datasets and tools were used in a manner consistent with their intended use. For instance, the Mol-Instructions dataset~\citep{fang2024molinstructions}—originally designed for molecule generation and property prediction—was employed for aligned downstream tasks in our study. Likewise, RDKit was used exclusively for molecular structure analysis and data preprocessing, as intended by its developers.

%% file: figure_appx/fig_appx_2_1_fg.tex
\begin{figure}[h]
  \centering
  \resizebox{0.9\linewidth}{!}{
  \begin{tcolorbox}[
    colback=gray!5!white, 
    colframe=gray!60!black,
    title={
      \parbox[t]{\dimexpr\linewidth-4mm\relax}{%
        \ttfamily
        Functional group matching

      }
    }
  ]

Answer only in 'Yes' or 'No' without any other information. \\

**Question:** Does the molecule represented by the SMILES string contain the specified functional group? Respond with 'Yes' or 'No'. \\
**SMILES:** \textcolor{blue}{[SMILES]} \\
**FUNCTIONAL GROUP:** \textcolor{blue}{[Functional group SMILES]} \\

**ANSWER:** \textcolor{blue}{[Yes/No]}

  \end{tcolorbox}
  }
    \caption{\textbf{An instruction format of functional group matching.} }\label{fig:appx_2_1_fg}
\end{figure}

%% file: figure_appx/fig_appx_2_2_ring.tex
\begin{figure}[h]
  \centering
  \resizebox{0.9\linewidth}{!}{
  \begin{tcolorbox}[
    colback=gray!5!white, 
    colframe=gray!60!black,
    title={
      \parbox[t]{\dimexpr\linewidth-4mm\relax}{%
        \ttfamily
        Ring counting

      }
    }
  ]
Answer only with the corresponding integer number without any other information. \\

**Question:** Assess the SMILES below and report how many rings consist of \textcolor{blue}{[RING SIZE]} atoms. Give me the integer only. \\
**SMILES:** \textcolor{blue}{[SMILES]} \\
**SIZE OF RINGS:** \textcolor{blue}{[RING SIZE]} \\

**ANSWER:** \textcolor{blue}{[NUMBER OF RINGS]}

  \end{tcolorbox}
  }
    \caption{\textbf{An instruction format of ring counting.} }\label{fig:appx_2_2_ring}
\end{figure}

%% file: figure_appx/fig_appx_2_3_chain.tex
\begin{figure}[h]
  \centering
  \resizebox{0.9\linewidth}{!}{
  \begin{tcolorbox}[
    colback=gray!5!white, 
    colframe=gray!60!black,
    title={
      \parbox[t]{\dimexpr\linewidth-4mm\relax}{%
        \ttfamily
        Chain length measurement

      }
    }
  ]

Answer only with the corresponding integer number without any other information. \\

**Question:** Report the size of the largest carbon-only chain not contained within a ring in the molecule represented by this SMILES. Answer with an integer only. \\
**SMILES:** \textcolor{blue}{[SMILES]} \\

**ANSWER:** \textcolor{blue}{[LENGTH OF CHAIN]}
  \end{tcolorbox}
  }
    \caption{\textbf{An instruction format of chain length measurement.} }\label{fig:appx_2_3_chain}
\end{figure}

%% file: figure_appx/fig_appx_2_4_cano.tex
\begin{figure}[h]
  \centering
  \resizebox{0.9\linewidth}{!}{
  \begin{tcolorbox}[
    colback=gray!5!white, 
    colframe=gray!60!black,
    title={
      \parbox[t]{\dimexpr\linewidth-4mm\relax}{%
        \ttfamily
        SMILES canonicalization

      }
    }
  ]

Answer only with the corresponding SMILES string without any other information. \\

**Question:** Give me a canonicalized SMILES string that represents the same molecule as the given one. \\

**SMILES:** \textcolor{blue}{[SMILES]}

**ANSWER:** \textcolor{blue}{[CANONICAL SMILES]}
  \end{tcolorbox}
  }
    \caption{\textbf{An instruction format of SMILES canonicalization.} }\label{fig:appx_2_4_cano}
\end{figure}

%% file: figure_appx/fig_appx_2_5_assemble.tex
\begin{figure}[h]
  \centering
  \resizebox{0.9\linewidth}{!}{
  \begin{tcolorbox}[
    colback=gray!5!white, 
    colframe=gray!60!black,
    title={
      \parbox[t]{\dimexpr\linewidth-4mm\relax}{%
        \ttfamily
        Fragment assembly

      }
    }
  ]

Answer only with the corresponding SMILES string without any other information. \\

**Question:** Connect the following two SMILES fragments into a unified structure at their reactive sites. \\
**SMILES:** \textcolor{blue}{[FRAGMENT 1, FRAGMENT 2]} \\

**ANSWER:** \textcolor{blue}{[SMILES]}
  \end{tcolorbox}
  }
    \caption{\textbf{An instruction format of SMILES assembly.} }\label{fig:appx_2_5_assemble}
\end{figure}

%% file: table_appx/tab_appx_1_hyper_parsing.tex
\begin{table}[h]
    \centering
    
    \resizebox{0.8\linewidth}{!}{
    \begin{tabular}{cc}
    \toprule
     & \textbf{Hyperparameter}  \\
    \midrule
    Batch size & 16 \\
    Learning rate & $5e^{-4}$ \\
    Epochs & 1 \\
    Warmup ratio & 0.01  \\
    Weight decay & 0.1  \\    
    Lr scheduler & cosine  \\
    Gradient accumulation steps & 1 \\
    Repetition penalty & 1 \\
    Temperature & 0.2 \\
    \midrule
    Lora r & 64 \\
    Lora alpha & 16 \\
    Lora dropout & 0.05 \\
    \bottomrule
    \end{tabular}}
    \caption{\textbf{Hyperparameters for SMILES parsing.}}
    \label{tab:appx_1_hyper_parsing}
\end{table}

%% file: table_appx/tab_appx_1_hyper_downstream.tex
\begin{table}[h]
    \centering
    
    \resizebox{0.8\linewidth}{!}{
    \begin{tabular}{cc}
    \toprule
     & \textbf{Hyperparameter}  \\
    \midrule
    Batch size & 16 \\
    Learning rate & $5e^{-4}$ \\
    Epochs & 1 \\
    Warmup ratio & 0.01  \\
    Weight decay & 0.1  \\    
    Lr scheduler & cosine  \\
    Gradient accumulation steps & 1 \\
    Repetition penalty & 1 \\
    Temperature & 0.2 \\
    \midrule
    Lora r & 64 \\
    Lora alpha & 16 \\
    Lora dropout & 0.05 \\
    \bottomrule
    \end{tabular}}
    \caption{\textbf{Hyperparameters for downstream tasks.}}
    \label{tab:appx_1_hyper_downstream}
\end{table}

%% file: table/tab_molecular_classification.tex
\begin{table}[h]
  \centering
  \resizebox{0.8\linewidth}{!}{
    \begin{tabular}{ccccc}
      \toprule[1.25pt]
      \textbf{Model} & \textbf{BACE} & \textbf{HIV} & \textbf{Clintox} \\
      \midrule[1.25pt]
    MolCA (1D+2D) & 0.798 & -- & 0.895 \\
    $\text{LlasMol}_{\text{Mistral}}$  & -- & 0.967 & 0.931 \\
    InstructMol-GS & 0.821 & 0.689 & -- \\
    \midrule
      LLaMA3.1-8B              & 0.507 & 0.971 & 0.946 \\
      \quad + \Algname          & \textcolor{teal}{0.639} & 0.971 & 0.946 \\
      \midrule
      Qwen2.5-7B                &  0.533  & 0.969 & 0.946 \\
      \quad + \Algname          & \textcolor{teal}{0.638} & \textcolor{teal}{0.971} & 0.946 \\
      \bottomrule[1.25pt]
    \end{tabular}
  }
  \caption{\textbf{Molecular property classification performance on the MoleculeNet dataset.}} 
  \label{tab:mol_classification}
\end{table}

%% file: table/tab_mol_reg.tex
\begin{table}[h]
  
  \centering
  \resizebox{0.8\linewidth}{!}{
    \begin{tabular}{cc}
      \toprule[1.25pt]
      \textbf{Model} & \textbf{MAE} \\
      \midrule[1.25pt]
    Alpaca & 322.109 \\
    Baize & 261.343 \\
    Vicuna & 860.051 \\
    Galactica &  \phantom{00}0.568 \\
    Mol-Instruct. (Lla.2) &  \phantom{00}0.013 \\
    Mol-Instruct. (Lla.3) & \phantom{0}15.059 \\
    Mol-Instruct. (Lla.3.1)* & \phantom{00}0.011 \\
    Mol-Instruct. (SFT)* & \phantom{00}\textbf{0.005} \\
    \midrule
      LLaMA3.1-8B               & \phantom{00}\textbf{0.005}  \\
      \quad + \Algname          & \phantom{00}\textbf{0.005} \\
      \midrule
      Qwen2.5-7B                 & \phantom{0}15.923 \\
      \quad + \Algname           & \phantom{00}\textbf{0.005} \\
      
      \bottomrule[1.25pt]
    \end{tabular}
  }
  \caption{\textbf{Molecular property regression performance on the Molinstructions dataset.}} 
  \label{tab:mol_reg}
\end{table}